\documentclass[10pt,twocolumn,letterpaper]{article}

\usepackage{wacv}              

\usepackage{graphicx}
\usepackage{amsmath}
\usepackage{amssymb}
\usepackage{booktabs}

\usepackage{algorithm}
\usepackage{algpseudocode}
\usepackage{float} %
\usepackage[accsupp]{axessibility} 

\usepackage[pagebackref,breaklinks,colorlinks]{hyperref}

\usepackage[capitalize]{cleveref}
\crefname{section}{Sec.}{Secs.}
\Crefname{section}{Section}{Sections}
\Crefname{table}{Table}{Tables}
\crefname{table}{Tab.}{Tabs.}

\def\wacvPaperID{1451} 
\def\confName{WACV}
\def\confYear{2025}

\begin{document}

\title{Learning Instance-Specific Parameters of Black-Box Models Using Differentiable Surrogates}

\author{Arnisha Khondaker \quad  Nilanjan Ray\\
University of Alberta, Canada\\
{\tt\small \{arnisha, nray1\}@ualberta.ca}}
\maketitle

\begin{abstract}
   Tuning parameters of a non-differentiable or black-box compute is challenging. Existing methods rely mostly on random sampling or grid sampling from the parameter space. Further, with all the current methods, it is not possible to supply any input specific parameters to the black-box. To the best of our knowledge, for the first time, we are able to learn input-specific parameters for a black box in this work. As a test application, we choose a popular image denoising method BM3D as our black-box compute. Then, we use a differentiable surrogate model (a neural network) to approximate the black-box behaviour. Next, another neural network is used in an end-to-end fashion to learn input instance-specific parameters for the black-box. Motivated by prior advances in surrogate-based optimization, we applied our method to the Smartphone Image Denoising Dataset (SIDD) and the Color Berkeley Segmentation Dataset (CBSD68) for image denoising. The results are compelling, demonstrating a significant increase in PSNR and a notable improvement in SSIM nearing 0.93. Experimental results underscore the effectiveness of our approach in achieving substantial improvements in both model performance and optimization efficiency. For code and implementation details, please refer to our GitHub repository: \href{https://github.com/arnisha-k/instance-specific-param} {https://github.com/arnisha-k/instance-specific-param}. 
\end{abstract}

\section{Introduction}
\label{sec:intro}

  Black-box models are systems or processes whose internal workings are unknown or inaccessible. These models are common across various domains; for instance, most Image Signal Processors (ISPs) in consumer-grade cameras, specialized imaging hardware, and simulators used for modeling intricate real-world scenarios or generating synthetic data for machine learning tasks function as black boxes  \cite{Ruiz2019,Richter2016,Ros2016,Behl2020}.

The configuration parameters of these black-box models often exhibit complex interactions with the output, playing a crucial role in determining efficacy and enhancing performance. Traditionally, manual tuning of these parameters has been labor-intensive, requiring extensive domain expertise. While effective, a manual method is neither scalable nor consistent. Automating the parameter tuning process can significantly reduce time and effort, leading to more efficient model configurations. 

Several methods exist for optimizing such black box models, depending on the availability of gradients \cite{lei2017review}. However, a challenge in optimization arises because many black-box models contain complex, non-differentiable components. Efficient non-differentiable black box optimization remains an open research problem with various solutions. For low-dimensional problems or scenarios where fast function evaluation is feasible, grid search remains a practical strategy\cite{Yu2020HyperParameterOA}. Early methods like Powell's bisection search \cite{Powell1965} and the Nelder-Mead method \cite{Nelder1965} have also been employed. Random search methods have been proposed for parameter optimization problems involving mixed categorical and continuous parameters \cite{Bergstra2012}, building probabilistic priors over function evaluations to approximate the objective landscape. Bayesian optimization methods \cite{Bergstra2013, Shahriari2016, Snoek2012, Swersky2013} differ in their surrogate modeling approaches but operate similarly in practice, though scaling to higher-dimensional parameter spaces remains challenging. Evolutionary algorithms \cite{papalambros2000principles,hansen2003reducing, Loshchilov2017} can scale to higher-dimensional spaces but require efficient objective evaluations, which limits their use with large training datasets. Stochastic gradient estimators \cite{mohamed2019monte} such as REINFORCE \cite{REINFORCE} have been utilized to estimate gradients for non-differentiable functions \cite{chen2016learning,stulp2013policy} and subsequently perform gradient-based optimization. In \cite{10.5555/2997046.2997074},the authors introduce a perceptron-like training algorithm that directly incorporates domain-specific loss functions—such as BLEU (BiLingual Evaluation Understudy) scores in machine translation—into the training process. Their approach provides a theoretical foundation for performing stochastic gradient descent directly on the loss of the inference system.

More recent approaches leverage surrogate or proxy models to approximate and optimize black box performance, mitigating the high variance associated with score function gradient estimators and efficiently solving high-dimensional parameter optimization problems \cite{TYY19, LGS}.

The Local Generative Surrogate Optimization (L-GSO) method is designed for gradient-based optimization of black-box simulators using differentiable local surrogate models \cite{LGS}. This approach is particularly beneficial in fields such as physics and engineering, where processes are often modeled with non-differentiable simulators that have intractable likelihoods. L-GSO employs deep generative models to iteratively approximate the simulator within local neighborhoods of the parameter space. This facilitates the approximation of the simulator’s gradient, enabling gradient-based optimization of the simulator parameters. L-GSO has proven efficient in high-dimensional parameter spaces, especially when the parameters reside on a low-dimensional submanifold. This approach demonstrates faster convergence to minima compared to baseline methods such as Bayesian optimization \cite{mockus1975bayesian}, numerical optimization \cite{nocedal2006numerical}, and score function gradient estimators \cite{williams1992simple}. 

In this context, Tseng et al. have pioneered the use of differentiable proxies for parameter optimization in black-box image processing \cite{TYY19}. Their method automates the parameter tuning process in both hardware and software image processing pipelines. By employing convolutional neural networks (CNNs) to generate differentiable proxies, they establish a mappable relationship between configuration parameters and evaluation metrics. This innovative technique enables efficient high-dimensional parameter search using stochastic first-order optimizers, eliminating the need to explicitly model the intricate lower-level transformations in image processing.
The authors demonstrate that by optimizing with this approach, traditional algorithms can outperform recent deep learning methods on specific benchmarks \cite{elad2006image, guo2018toward, dabov2007image}.

While existing methods demonstrate the capability to tune black-box parameters, \textbf{they lack the ability to deliver input-specific parameter tuning}. Inspired by Tseng et al.'s work \cite{TYY19}, this work automates input-specific black-box model parameter optimization using a differentiable surrogate. Unlike \cite{TYY19}, we adopt UNETR architecture \cite{hatamizadeh2021unetr} as our surrogate CNN to model the black box and extend the method with the following contributions:

\begin{enumerate}
    \item End-to-end training with the black-box  in the loop to dynamically update its parameters to generate training instances, thereby improving the surrogate approximator's adaptivity and performance.
    \item Using a predictive model to learn black-box parameters specific to individual inputs.
    \item Demonstrating improved image denoising performance on the SIDD  \cite{Abdelhamed2018DenoisingDataset} and CBSD68 \cite{MartinFTM01} dataset using the BM3D denoiser \cite{dabov2007image} compared to other proxy-based optimization methods.
\end{enumerate}

 The rest of this paper is organized as follows: \Cref{sec2:background} discusses a brief background of the foundational concepts drawn from existing literature relevant to our work, while \Cref{sec3:proposed_method} presents our approach. \Cref{sec4:exps} details the implementation setup and results, and \Cref{sec5:conclusion} concludes with future work.

\section{Background}
\label{sec2:background}
\subsection{Black-Box Approximation}
The work of Randika et al. aims to optimize the parameters of an Optical Character Recognition (OCR) preprocessor by approximating the gradient of the OCR engine, treating it as a black-box due to the inaccessibility of its internal mechanisms and the potential presence of non-differentiable components \cite{Randika2021UnknownBox}. It stands in contrast to previous OCR-agnostic preprocessing techniques, such as Otsu's binarization \cite{Otsu1979Threshold}, skeletonization \cite{LamSuen1995ParallelThinning}, geometrical correction methods \cite{Bieniecki2007ImagePreprocessing}, and deep learning-based methods like DeepOtsu \cite{HeSchomaker2019DeepOtsu}.

The approach involves an alternating optimization process, where noise is added to the input in an inner loop to perturb the OCR engine and accumulate errors between the OCR output and an approximating neural network. This neural network is trained to approximate the OCR’s gradient, enabling the optimization of the preprocessor parameters. The algorithm alternates between minimizing the error of the preprocessor and the approximating neural network, thereby enhancing OCR performance through effective gradient approximation and parameter tuning for the preprocessor. This method demonstrates improved OCR accuracy on challenging datasets \cite{huang2019icdar,jaderberg2014synthetic}.

Our present setup uses the same alternating differentiable approximating mechanism. While Randika et al. \cite{Randika2021UnknownBox} uses a CRNN model for approximation, we use a UNETR architecture.

\subsection{Black-Box Parameter Optimization}

In optimizing black-box imaging systems' parameters using differentiable proxies, Tseng et al.\cite{TYY19} propose Algorithm \ref{algo1}. First, a Differentiable Proxy Function (modelled with a CNN) is trained to emulate the behavior of a given Image Signal Processor (ISP) on input images with its parameters. This training phase minimizes loss between proxy's output and the actual ISP output across multiple images and parameter samples. Then, the optimized weights obtained from the first stage are applied to instantiate a relaxed parameter optimization problem which involves minimizing a task-specific loss function over the ISP parameters, utilizing the proxy's partial derivatives with respect to the parameters. This framework allows for gradient-based optimization methods, enhancing scalability across larger parameter spaces and nonlinearities in black-box ISP systems. As we build upon these insights, concepts derived from \cite{TYY19} for our method are explained in Sections \ref{subsec:2.3} and \ref{subsec:2.4}.

\subsection{Modeling the Black-Box System}
\label{subsec:2.3}
The black-box model is conceptualized as abstract function \( f_{\text{bb}} \) that map an input $x$ to an output $y$. These models are parameterized by a set of parameters $\theta$, such that $y = f_{bb}(x;\theta)$. Our objective  is to address the parameter optimization problem, which can be formulated as follows:
\begin{equation}
\label{eq_bb}
\theta^* = \underset{\theta}{\operatorname{arg\,min}} \sum_{i=1}^{N} \mathcal{L} \left( f_{\text{bb}}(x_i; \theta), y_i \right)
\end{equation}
where $\theta^* $ represents the optimal set of parameters, $x_i$  the $i$-th input instance, $y_i$ the corresponding target output, $f_{\text{bb}}(x_i; \theta)$ the predicted output, $\mathcal{L} \left( \cdot, \cdot \right)$ the task-specific loss function, and $N$ the number of target instances.

Further, we want to learn an instance-specific parameter learner $\theta_i = f_{\text{pl}}(x_i;\psi)$  parameterized by $\psi$:
\begin{equation}
\label{eq_instance}
\psi^* = \underset{\psi}{\operatorname{arg\,min}} \sum_{i=1}^{N} \mathcal{L} \left( f_{\text{bb}}(x_i; f_{\text{pl}}(x_i;\psi)), y_i \right)
\end{equation}
with the hope that the instance-specific learning of black-box parameters would be more effective.

\subsection{Training an Approximator Model to Learn the Black-Box Function}
\label{subsec:2.4}
Typically, the function $f_{\text{bb}}$ of the black box model is unknown or non-differentiable, complicating the solution of Eq. (\ref{eq_bb}) and (\ref{eq_instance}). This challenge can be addressed using a differentiable approximation  $f_{\text{am}}$ for $f_{\text{bb}}$. The approximation model, similar to $f_{\text{bb}}$, maps inputs to outputs and also accepts parameters $\theta$ as inputs. $f_{\text{am}}$ is parameterized by its weights $\omega$ and ensures it is differentiable with respect to $\theta$ and $\omega$. The optimization problem for the approximator model is:
\begin{equation}
\label{eq_proxropt}
\omega^* = \underset{\omega}{\operatorname{arg\,min}} \sum_{i=1}^{N}\underset{\theta}{\mathbb{E}}~\mathcal{L}( f_{\text{am}}(x_i, \theta; \omega), f_{\text{bb}}(x_i; \theta))
\end{equation}
where the loss $\mathcal{L}$ ensures that the approximator closely resembles the output of the black-box model. Since  $f_{\text{am}}$ is differentiable with respect to $\omega$, Tseng et al. \cite{TYY19} used first-order techniques to mimic black-box outputs and obtain optimized weights $\omega^*$ (see Algorithm \ref{algo1}) . This optimization simplifies the black-box parameter optimization problem to:
\begin{equation}
\label{eq_proxropt2}
\theta^* = \underset{\theta}{\operatorname{arg\,min}} \sum_{i=1}^{N} \mathcal{L}(f_{\text{am}}(x_i, \theta; \omega^*), y_i)
\end{equation}
Here, substituting $f_{\text{bb}}$ with $f_{\text{am}}$ from Eq. (\ref{eq_bb}) transforms the problem into a more tractable form, allowing evaluation of $f_{\text{am}}$ and its gradient with respect to the parameters. This allows the application of first-order optimization methods as shown in Algorithm (\ref{algo1}). Note that in Algorithm (\ref{algo1}) Tseng et al. \cite{TYY19} needed to create an instance of parameter vector $\theta_i$ for every input instance $x_i$. This was done by randomly sampling $\theta$. We implemented Algorithm (\ref{algo1}) using mini-batches from the training set.

\begin{algorithm}
    \caption{Black-Box Optimization}
    \label{algo1}
    \begin{algorithmic}
        \Require Dataset $D = \{(x_i, y_i, \theta_i)\}_{i=1}^N$, initial surrogate model weights $\omega$, number of iterations $\tau$, learning rate $\alpha > 0$
        \Ensure Surrogate $f_{\text{am}}$ with optimized weights $\omega^*$ and optimum black-box parameters $\theta^*$
        \For {$t \gets 0$ \textbf{to} $\tau - 1$}
                \State $L_a \gets \sum_i \mathcal{L}(f_{\text{am}}(x_i, \theta; \omega), f_{\text{bb}}(x_i; \theta))$
                \State $\omega \gets \omega - \alpha \nabla_\omega L_a$
        \EndFor
        \State Initialize $\theta$
        \For {$t \gets 0$ \textbf{to} $\tau - 1$}
                \State $L_s \gets \sum_i \mathcal{L}(f_{\text{am}}(x_i, \theta; \omega), y_i)$
                \State $\theta \gets \theta - \alpha \nabla_{\theta} L_s$
        \EndFor
        \State \textbf{return} Learned Black-Box parameters $\theta$
    \end{algorithmic}
\end{algorithm}

\section{Proposed Method }
\label{sec3:proposed_method}
\begin{figure*}[t]
\begin{center}
\includegraphics[width=0.90\linewidth]{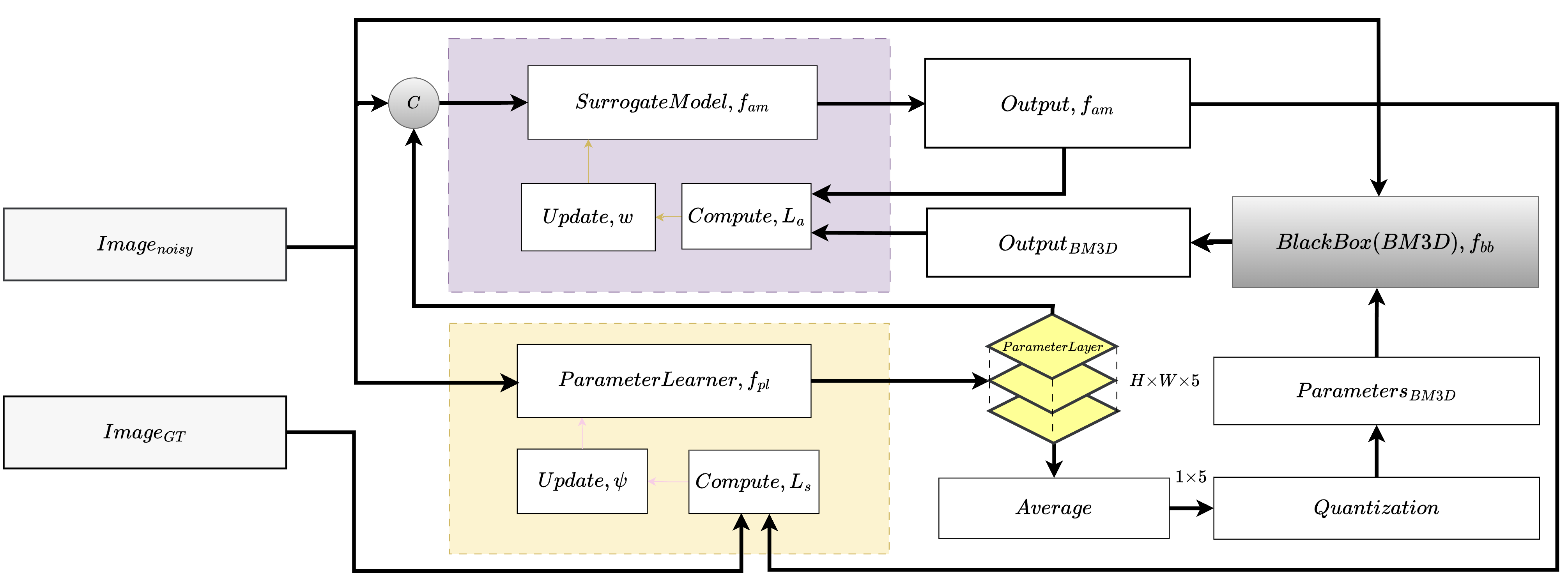}
\end{center}
   \caption{Illustration of instance-specific parameter learning in BM3D denoising application (Algorithm \ref{algo3}). The surrogate model \(f_{\text{am}}\) takes the noisy image and parameters as input channels (\Cref{fig:proxy_unetr}). The loss \(\omega\) is updated using the model output and BM3D output. The parameter learner \(f_{\text{pl}}\) outputs a parameter layer, which in the case of BM3D is \(H \times W \times 5\). This layer is averaged across each channel and quantized back into the parameter space to obtain the BM3D output with updated parameters. The input to the parameter learner is noisy images, and \(\psi\) is updated using the approximator output and ground truth images.}
\label{fig:training}
\end{figure*}
\subsection{Dynamic Optimization with Differentiable Surrogate}

\begin{algorithm}
    \caption{Dynamic Black-Box Optimization}
    \label{algo2}
    \begin{algorithmic}
        \Require Dataset $D = \{(x_i, y_i, \theta_i)\}_{i=1}^N$, initial surrogate model weights $\omega$, number of iterations $\tau$, learning rate $\alpha > 0$
        \Ensure Surrogate $f_{\text{am}}$ with optimized weights $\omega^*$ and optimum black-box parameters $\theta^*$        
        \For {$t \gets 0$ \textbf{to} $\tau - 1$}
                \State $L_a \gets \sum_i \mathcal{L}(f_{\text{am}}(x_i, \theta; \omega), f_{\text{bb}}(x_i; \theta))$
                \State $\omega \gets \omega - \alpha \nabla_\omega L_a$
                \State $L_s \gets \sum_i \mathcal{L}(f_{\text{am}}(x_i, \theta; \omega), y_i)$
                \State $\theta \gets \theta - \alpha \nabla_{\theta} L_s$
        \EndFor
        \State \textbf{return} Learned Black-Box parameters $\theta$
    \end{algorithmic}
\end{algorithm}


Algorithm (\ref{algo1}) uses a differentiable surrogate function $f_{\text{am}}$ to model $f_{\text{bb}}$ using randomly sampled parameters of the black-box. Here we propose a dynamic variation of Algorithm (\ref{algo1}) following Ayantha et al. \cite{Randika2021UnknownBox} (see Algorithm \ref{algo2}). The important difference between Algorithm (\ref{algo1}) and (\ref{algo2}) is that in every iteration the surrogate function dynamically approximates the black-box and simultaneously updates the black-box parameter vector. As in Algorithm (\ref{algo1}), we used mini-batching to implement gradient-based optimization, e.g., the Adam optimizer. 
Gradient update with mini-batch for $\omega$ is followed by gradient mini-batch update of $\theta$. For the image denosing application, we demonstrate that the dynamic version (Algorithm \ref{algo2}) yields significantly better accuracy for BM3D with more effective parameter tuning. Note also that we used proper quantization of continuous $\theta$ before passing to $f_{\text{bb}}$ when needed, but we pass continuous $\theta$ to the approximator $f_{\text{am}}$.

\subsection{Learning Parameters of Black-Box}

\begin{algorithm}
    \caption{Learning Black-box Parameters}
    \label{algo3}
    \begin{algorithmic}
        \Require Dataset $D = \{(x_i, y_i)\}_{i=1}^N$, initial surrogate model weights $\omega$, initial parameter learner weights $\psi$, number of iterations $\tau$, learning rate $\alpha > 0$
        \Ensure Surrogate $f_{\text{am}}$ with optimized weights $\omega^*$ and parameter learner $f_{\text{pl}}$ with optimized weights $\psi^*$
        \For {$t \gets 0$ \textbf{to} $\tau - 1$}
            \State $\theta_i \gets f_{\text{pl}}(x_i;\psi), ~\forall i$
            \State $L_a \gets \sum_i \mathcal{L}(f_{\text{am}}(x_i,\theta_i; \omega), f_{\text{bb}}(x_i; \theta_i))$
            \State $\omega \gets \omega - \alpha \nabla_\omega L_a$
            \State $L_s \gets \sum_i \mathcal{L}(f_{\text{am}}(x_i, f_{\text{pl}}(x_i;\psi); \omega), y_i)$
            \State $\psi \gets \psi - \alpha \nabla_{\psi} L_s$
        \EndFor
        \State \textbf{return} Parameter learner $f_{\text{pl}}$
    \end{algorithmic}
\end{algorithm}

        
            
            

Futher to Algorithm (\ref{algo2}), we integrate an additional model, a parameter learner $f_{\text{pl}}$ to learn input instance-specific black-box parameters (see Eq. (\ref{eq_instance})). This model takes as input the current instance $x$ and learns to predict instance-specific optimal parameters that maximize the performance of the black-box model $f_{\text{bb}}$ for that specific instance. This is illustrated in Algorithm (\ref{algo3}). Our experiments show that learning of instance-specific black-box parameter further enhances the accuracy of BM3D.

\section{Experimental Setup and Results}
\label{sec4:exps}

\subsection{Dataset}


We employed the Smartphone Image Denoising Dataset (SIDD) developed by Abdelhamed et al. \cite{Abdelhamed2018DenoisingDataset} and the CBSD68 dataset from the Berkeley Segmentation Dataset and Benchmark\cite{MartinFTM01} in our experiments. Specifically, for SIDD, we focused on 23 scene instances captured using an iPhone 7 under normal lighting conditions, with ISO settings greater than 200. From these instances, we extracted 150 random crops of size 512 $\times$ 512 pixels, resulting in a total of 6,900 images. This subset included 3,450 pairs of ground truth (GT) and noisy images. For training and validation, we used 5,514 images (2,757 GT-NOISY pairs), where the training set comprised 2,413 image pairs (approximately 87.5\%), and the validation set consisted of 344 image pairs (approximately 12.5\%). To evaluate the denoising performance, 1,386 images (693 GT-NOISY pairs) were utilized for testing.

In addition, we employed the CBSD68 dataset, commonly used for image denoising benchmarks. This dataset introduces noise with a sigma ($\sigma$) value of 15. For training, we utilized 94 instances, including 47 GT-NOISY pairs. For validation, 12 instances (6 GT-NOISY pairs) and for testing, 15 GT-NOISY pairs were used.

\begin{figure}[t]
\begin{center}
    \begin{subfigure}[b]{0.3\linewidth}
        \includegraphics[width=\linewidth]{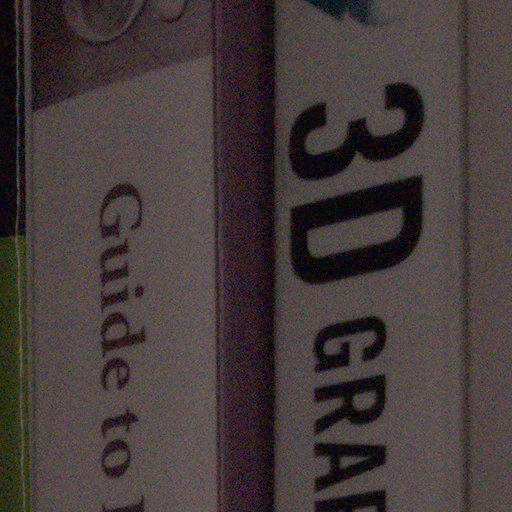}
        \caption{}
        \label{fig:image1}
    \end{subfigure}\hfill
    \begin{subfigure}[b]{0.3\linewidth}
        \includegraphics[width=\linewidth]{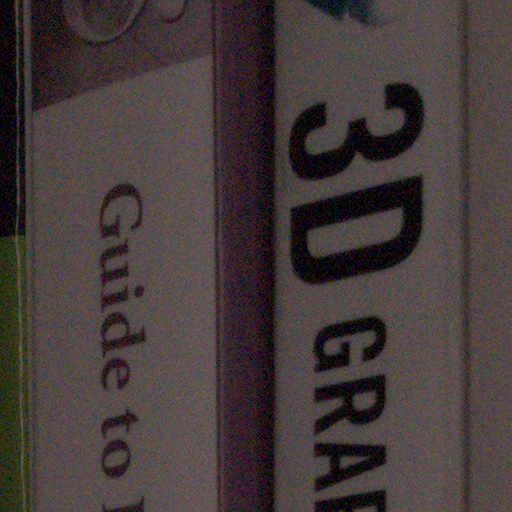}
        \caption{}
        \label{fig:image2}
    \end{subfigure}\hfill
    \begin{subfigure}[b]{0.3\linewidth}
        \includegraphics[width=\linewidth]{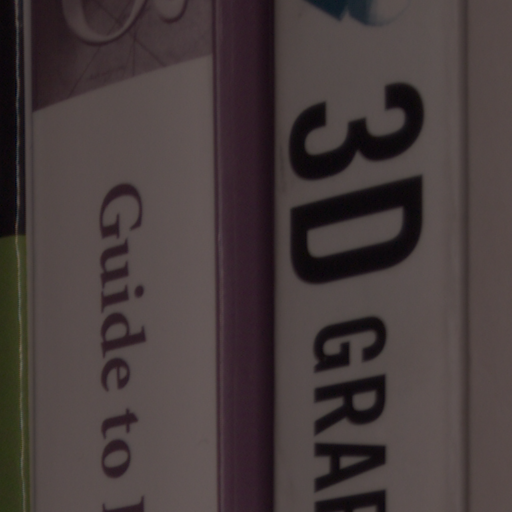}
        \caption{}
        \label{fig:image3}
    \end{subfigure}
\end{center}
\caption{Each sample of data comprises three key elements: (a) the initial unprocessed noisy image, (b) the image processed using the BM3D algorithm with specified parameters, and (c) the corresponding ground truth image with low noise.}
\label{fig:data}

\end{figure}

The dataset components are as follows (see \Cref{fig:data}):
\begin{itemize}
    \item \textbf{Unprocessed Input Images ({$ I$}):} Raw noisy images serving as inputs.
    \item \textbf{BM3D Output with Parameters ($y_{\text{BM3D}} + \theta$):} Images denoised by  BM3D after applying parameters $\theta$.
    \item \textbf{Ground Truth Images ($I^\text{gt}$):} Corresponding denoised ground truth images for each raw input.
\end{itemize}

\subsection{Evaluation Metrics}
We evaluated our proposed method using Peak Signal-to-Noise Ratio (PSNR)\cite{psnr1974} and Structural Similarity Index Measure (SSIM)\cite{ssim2004}, metrics for measuring the quality of an image or signal. Higher values of PSNR and SSIM indicate better image quality.








We compared our proposed methods with Algorithm (\ref{algo1}) as well as with random and exhaustive search to do a thorough performance test. The PSNR and SSIM values for each method are summarized in Table (\ref{tab:results-psnr}) and (\ref{tab:ssim_results}).


\subsection{Black Box: BM3D}

In our experiments, we utilize the BM3D denoiser—a 3D block-matching algorithm known for effectively reducing noise in images—as our black box model \cite{dabov2007image}. Consequently, Eq. (\ref{eq_instance})
is formulated as:
\begin{equation}
\label{eq_instance_BM3D}
\psi^* = \underset{\psi}{\operatorname{arg\,min}} \sum_{i=1}^{N} \mathcal{L} \left( f_{\text{BM3D}}(I_i; f_{\text{pl}}(I_i;\psi)), I^{\text{gt}}_i \right)
\end{equation}

Here, $f_{\text{BM3D}}(I_i; f_{\text{pl}}(I_i;\psi))$ denotes the output of the BM3D denoiser applied to input image $I_i$ with predicted parameters, and $I_i^{\text{gt}}$ denotes the corresponding ground truth clean image.

 The key tuning parameters (see Table \ref{tab:bm3d_hyperparameters}) of BM3D include patch size ($n1$), color space ($cspace$), Wiener transform type ($wtransform$), linear scale for noise ($cff$), and neighborhood size ($neighborhood$). The parameters \( cspace \) and \( wtransform \) are categorical variables. Specifically, for \( cspace \), the value 0 corresponds to the ``opp'' color space, and 1 corresponds to ``YCbCr''. For \( wtransform \), the value 0 represents the Discrete Cosine Transform (``dct''), and 1 represents the Discrete Sine Transform (``dst'').

\begin{table}[h!]
  \begin{center}
    {\small{
    \begin{tabular}{ll}
    \toprule
    \textbf{Parameter Name} & \textbf{Search Range} \\
    \midrule
    cff & Continuous in [1, 20] \\
    n1 & Discrete in \{ 4, 8\} \\
    cspace& Discrete in \{0, 1\} \\
    wtransform & Discrete in \{0, 1\} \\
    neighborhood & Discrete in \{ 3, 4, 5, \ldots, 15 \} \\
    \bottomrule
    \end{tabular}
    }}
  \end{center}
  \caption{BM3D Parameters Search Ranges}
  \label{tab:bm3d_hyperparameters}
\end{table}
\subsection{Algoritm 1: Black-Box Optimization}

\begin{figure}[t]
\begin{center}
    \includegraphics[width=\linewidth]{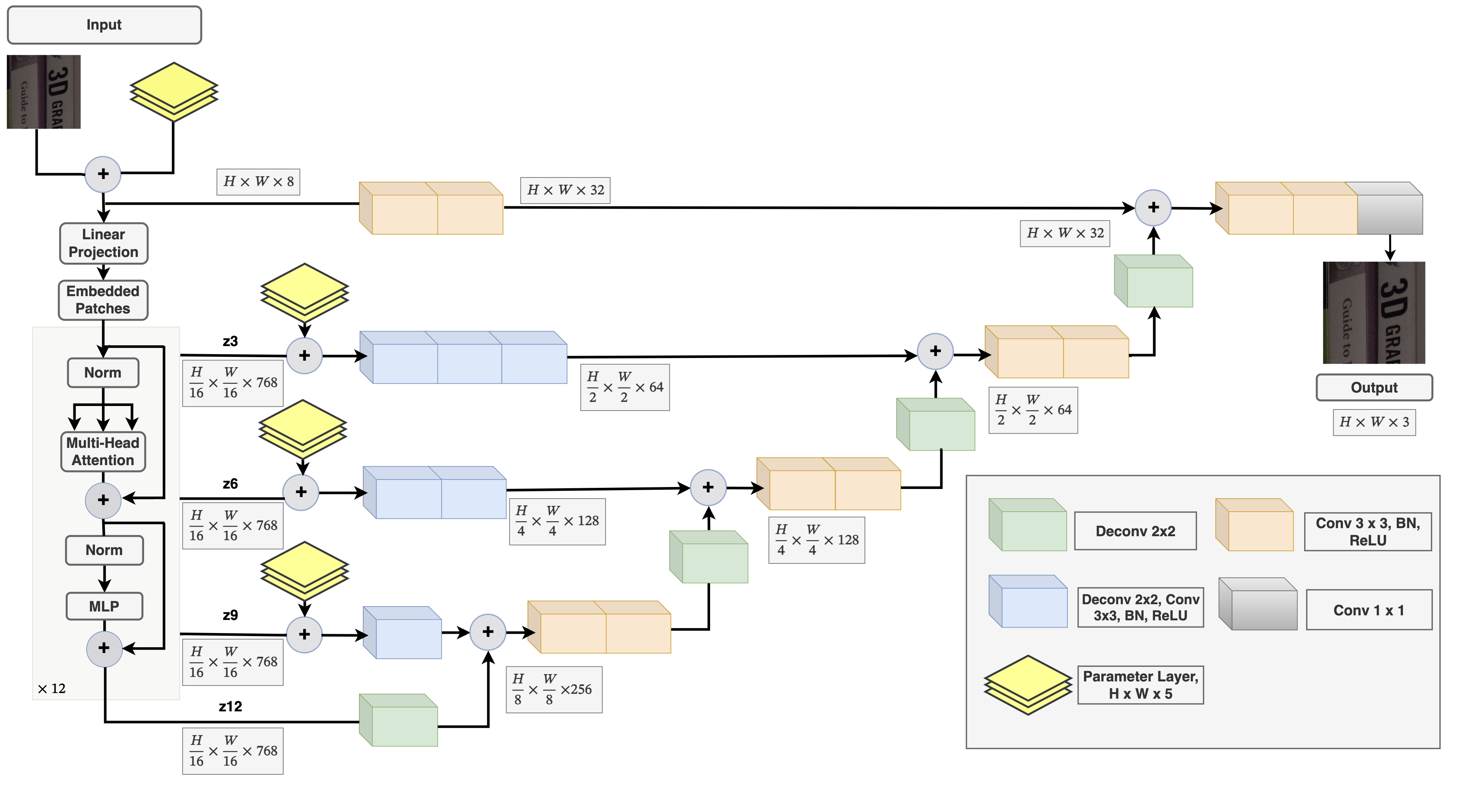}
\end{center}
   \caption{Architecture of the Surrogate Model($f_{\text{am}}$), UNETR. Output sizes are given for patch resolution = 16, filters = [32 , 64, 128, 256], attention heads = 12 and embedding size = 78. Parameter layer is concatenated with input as well as hidden layers 3, 6 and 9.}
\label{fig:proxy_unetr}
\end{figure}

For Algorithm (\ref{algo1}), we utilize UNETR (U-Net Transformers) (Figure \ref{fig:proxy_unetr}) to train a differentiable surrogate model instead of U-Net \cite{Ronneberger2015UNet} used in the original work for fair comparison with our method. UNETR integrates the Vision Transformer (ViT) \cite{Dosovitskiy2020ImageWorth} with a U-Net architecture \cite{Ronneberger2015UNet}, combining a transformer-based encoder with a U-shaped architecture that excels at learning sequence representations from input volumes and capturing essential global and multi-scale information \cite{hatamizadeh2021unetr}. 

To minimize weights $\omega$ of the surrogate model, we utilize two components from the dataset: unprocessed images $I$ and outputs $y_{\text{BM3D}}$ from BM3D  along with its parameters $\theta$. During training, these inputs are fed into UNETR to approximate BM3D's output and functional behavior. The approximation stage uses an eight-channel input composed of three image channels and five parameters (treated as additional channels). This input is processed by the ViT to extract global features using self-attention mechanisms. The model then transitions into a U-Net-like structure with encoder and decoder layers.

\begin{figure}[h]
\centering
    \begin{subfigure}[b]{0.5\linewidth}
        \includegraphics[width=\linewidth]{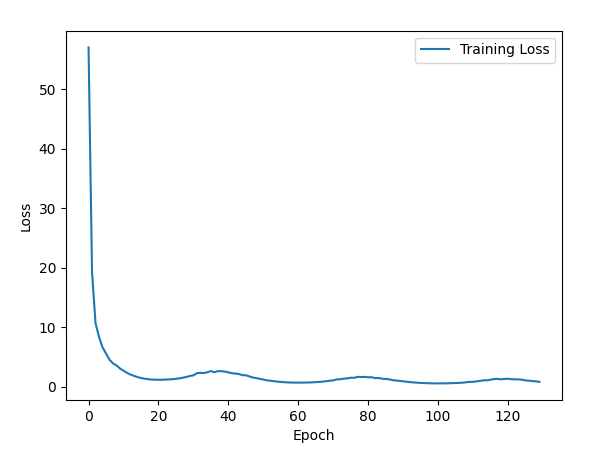}
        \caption{}
        \label{fig:image1}
    \end{subfigure}\hfill
    \begin{subfigure}[b]{0.48\linewidth}
        \includegraphics[width=\linewidth]{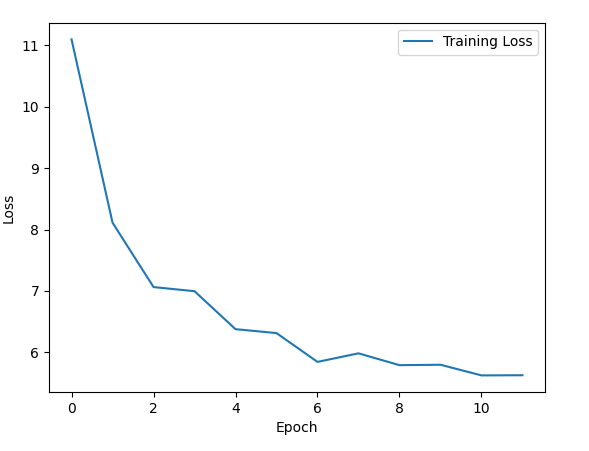}
        \caption{}
        \label{fig:image2}
    \end{subfigure}
\caption{Illustration of the losses during Algorithm \ref{algo1} (a) shows the surrogate approximator loss, whereas (b) shows the parameter optimization loss over epochs.}
\label{fig:algo1loss}
\end{figure}

Encoders' filter sizes increase from 32 to 64, 128, and 256, while the decoders reverse this operation. Transformer hidden states 3, 6 and 9 are integrated at the first 3 stages of the encoder-decoder structure (see Figure \ref{fig:proxy_unetr}). Before passing them onto the encoders, parameter channels are concatenated with these hidden states. The model is trained with a learning rate of 0.005 using ADAM optimizer and a batch size of 1, achieving convergence around epoch 20 (\Cref{fig:algo1loss}). 

Upon obtaining the approximator UNETR, the 5-channel parameter layer is optimized using ADAM optimizer with a Mean Squared Error (MSE) loss function. The initial learning rate is set to 0.02 and is reduced every 2 epochs according to the schedule \( \text{LR}_{i+1} = 0.8 \times \text{LR}_i \). The parameters are adjusted until convergence, which typically occurs rapidly, usually within 8 epochs (\Cref{fig:algo1loss}). 

\subsection{Algoritm 2:  Dynamic Black-Box Optimization}
\begin{figure}[h]
\centering
    \begin{subfigure}[b]{0.5\linewidth}
        \includegraphics[width=\linewidth]{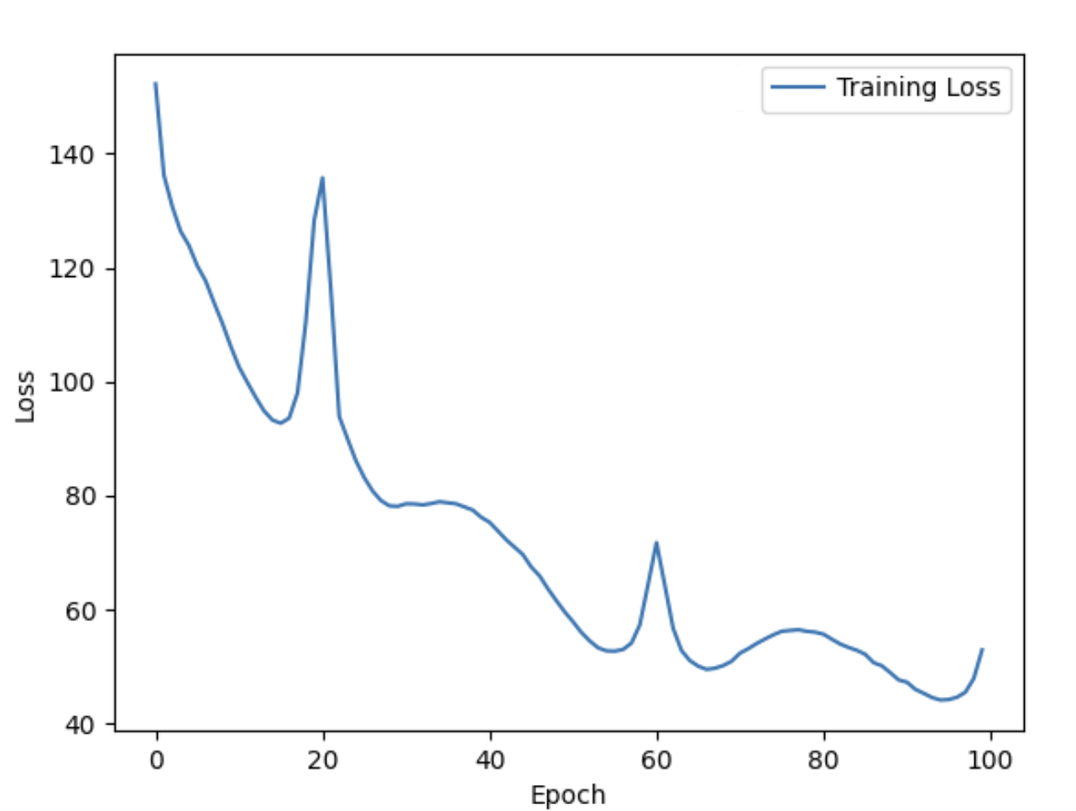}
        \caption{}
        \label{fig:image1}
    \end{subfigure}\hfill
    \begin{subfigure}[b]{0.48\linewidth}
        \includegraphics[width=\linewidth]{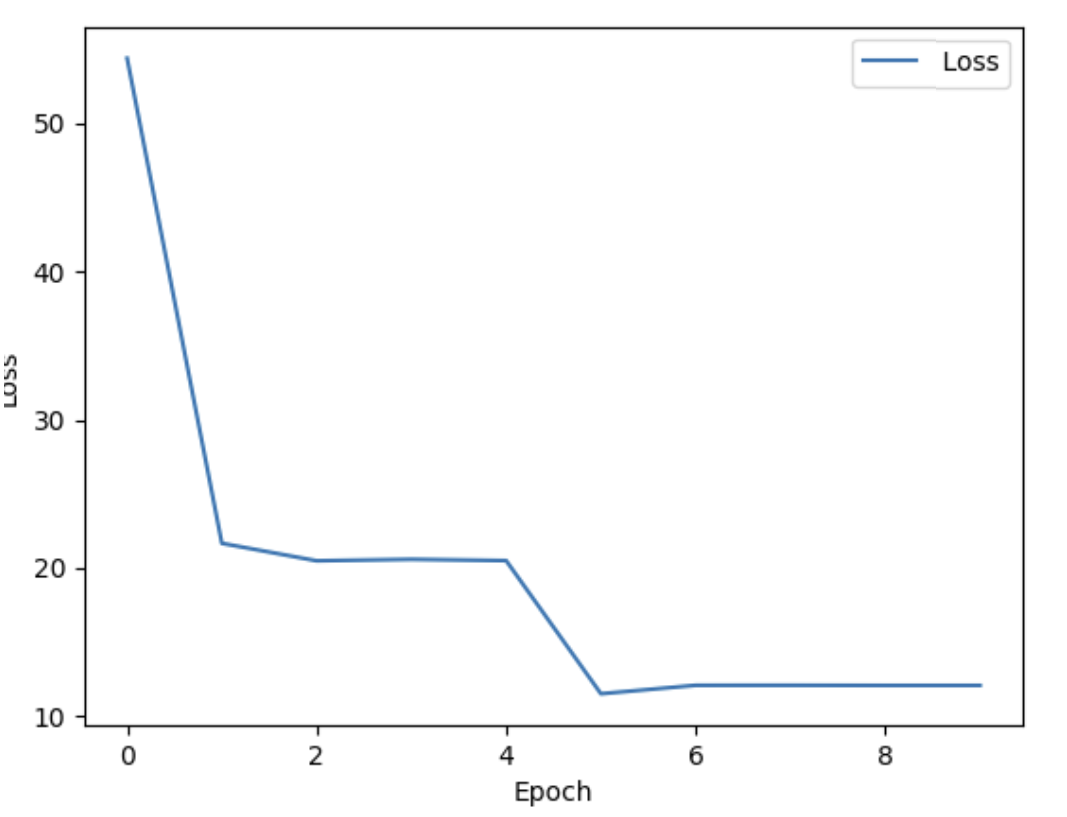}
        \caption{}
        \label{fig:image2}
    \end{subfigure}
\caption{Illustration of the losses during Algorithm \ref{algo2} (a) shows the surrogate approximator loss, whereas (b) shows the parameter optimization loss over epochs.}
\label{fig:algo2loss}
\end{figure}

For Algorithm (\ref{algo2}), we also employ UNETR  to train a differentiable surrogate model, similar to Algorithm (\ref{algo1}).  The surrogate loss $L_a$ is computed based on the BM3D outputs given current optimized parameters, and the surrogate model weights $\omega$ are updated accordingly. Once a sufficiently accurate approximation of the black-box model is achieved, typically around 10-20 iterations of training, we proceed to optimize the parameters $\theta$. The parameters are optimized with ADAM optimizer using MSE loss and a learning rate of 0.02. After optimization, the tuned parameter layer is averaged across each channel and quantized for input into BM3D. Subsequently, the BM3D processed images with the updated $\theta$ are used for calculating the loss of the approximator for next iteration, maintaining an end-to-end training approach. The updated non-quantized parameter layer is concatenated with the input and hidden states 3, 6 and 9 for subsequent iterations. The convergence plot for this training is illustrated in Figure \ref{fig:algo2loss}.

\subsection{Algorithm 3: Learning Black-Box Parameters}

In addition to using UNETR for approximation, in Algorithm (\ref{algo3}) we use a U-Net 
to predict the optimal set of parameters 
\({\theta_i}\) for each input image, $I_i$. The U-Net model is trained using Mean Squared Error (MSE) loss and optimized with the Adam optimizer, employing a learning rate set at 0.02. Our U-Net architecture initiates with convolutional layers featuring 64 filters, progressively scaling up to 512 filters 
following the original design of U-Net \cite{Ronneberger2015UNet}. The input to the parameter learner, $f_{\text{pl}}$ is the noisy image of dimension \( H \times W \times 3 \), and it produces an output of dimension \( H \times W \times 5 \). This learner is integrated into the training loop in an end-to-end fashion, as depicted in \Cref{fig:training}. The parameter processing is analogous to Algorithm (\ref{algo2}), and the surrogate model is also trained similarly.

\begin{figure}[h!]
\centering
    \begin{subfigure}[b]{0.5\linewidth}
        \includegraphics[width=\linewidth]{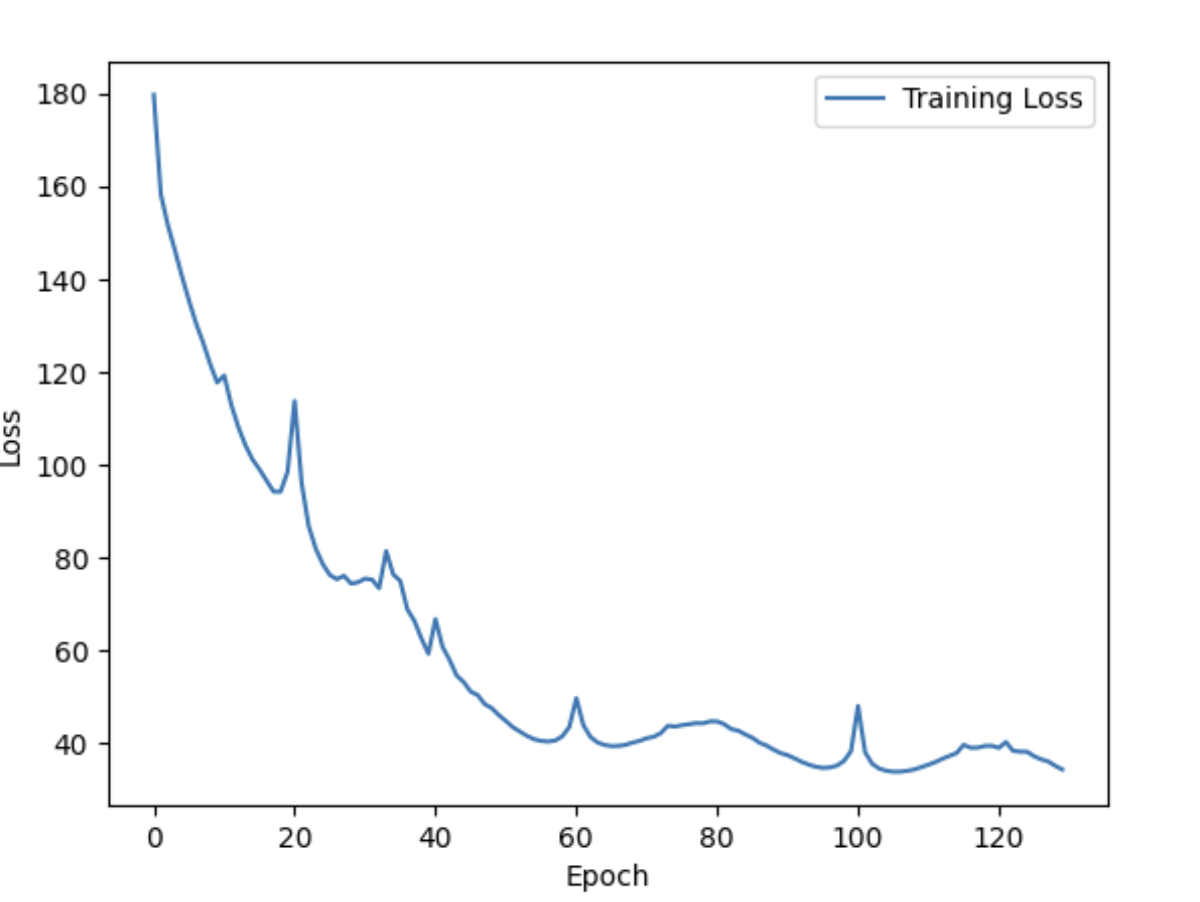}
        \caption{}
        \label{fig:image1}
    \end{subfigure}\hfill
    \begin{subfigure}[b]{0.48\linewidth}
        \includegraphics[width=\linewidth]{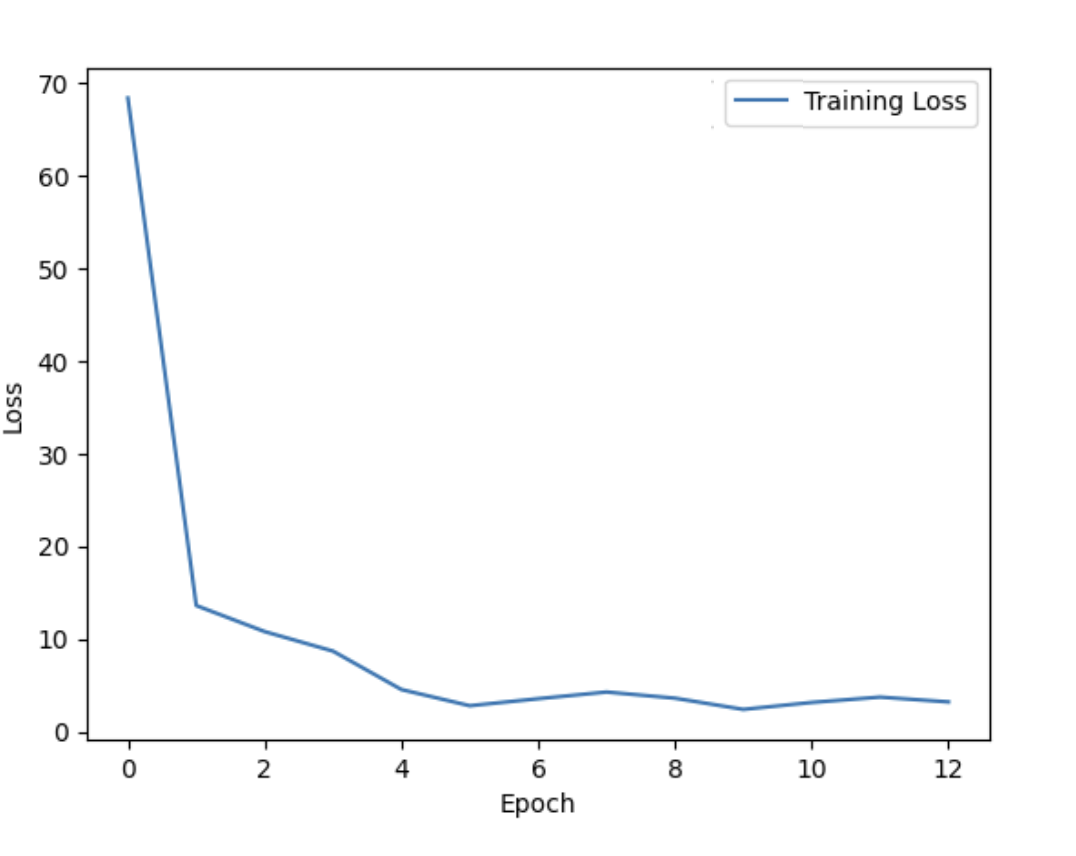}
        \caption{}
        \label{fig:image2}
    \end{subfigure}
\caption{Illustration of the losses during Algorithm \ref{algo3} (a) shows the approximation loss, whereas (b) shows the parameter learner optimization loss over epochs.}
\label{fig:algo3-loss}

\end{figure}

The approximation loss typically converges around epoch 60, while the optimization loss stabilizes by epoch 4. Figure \ref{fig:algo3-loss} shows detailed convergence trajectories of both the approximation and optimization losses for \Cref{algo3}.

\begin{figure}[h]
\begin{center}
    \includegraphics[width=\linewidth]{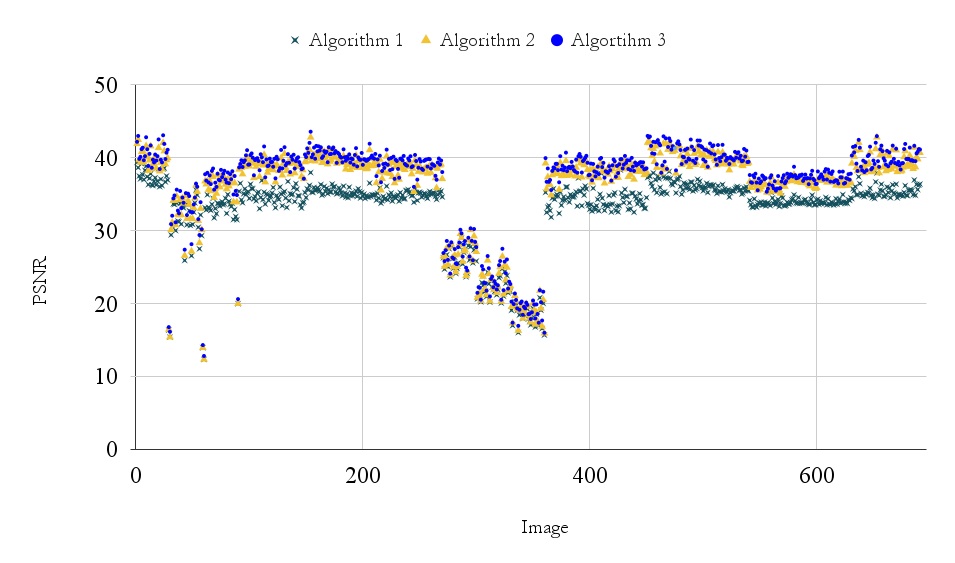}
\end{center}
   \caption{Scatter plot of \Cref{algo1}, \Cref{algo2}, and \Cref{algo3} performance on the test samples based on PSNR per image.}
\label{fig:scatter}
\end{figure}
\subsection{Results}



Our experiments were conducted on a system equipped with an NVIDIA GeForce RTX 4090 GPU, offering 24 GiB of memory. We explored baseline methods such as exhaustive and random search. For the exhaustive search, we evenly sampled the continuous parameter ($cff$) and treated other parameters as discrete. Due to the exponential complexity of the problem, particularly with slow BM3D calls, completing this experiment would take several months given the large parameter space and number of images. Random search proved feasible by randomly sampling from the parameter space and iteratively optimizing parameters based on the PSNR metric through BM3D. Each set of 10 iterations in random search took approximately 72 hours. Algorithms \ref{algo1}, \ref{algo2}, and \ref{algo3} were run for the same duration. During inference, we use BM3D with optimized parameters.

\begin{table}[h]
  \begin{center}
    \small{
      \begin{tabular}{lrr}
        \toprule
        \multicolumn{3}{c}{\textbf{Smartphone Image Denoising Dataset (SIDD) \cite{Abdelhamed2018DenoisingDataset}}} \\
        \midrule
        \textbf{Method} & \textbf{Train} & \textbf{Test} \\
        
        Proxy-Opt. (Alg. \ref{algo1}) & 33.41 & 33.23(±4.14)\\
        Random Grid Search  & 36.19 &  36.00 (±6.2)\\
        Dynamic-Opt. (Alg. \ref{algo2}) & 36.31$\uparrow$ & 36.12$\uparrow$(±6.0) \\
        Inst-Spec-Opt.(\(\mu\)) (Alg. \ref{algo3}) & 36.80$\uparrow$ & 36.51$\uparrow$(±5.8) \\
        Inst-Spec-Opt.(\(med\)) (Alg. \ref{algo3}) & 36.21$\uparrow$ & 36.11$\uparrow$(±5.9) \\
        \toprule
        \multicolumn{3}{c}{\textbf{Color Berkeley Segmentation Dataset (CBSD68) \cite{MartinFTM01}}} \\
        \midrule
        \textbf{Method} & \textbf{Train} & \textbf{Test} \\
        Proxy-Opt. (Alg. \ref{algo1}) & 31.12 & 30.91 (±1.9)\\
        Random Grid Search  & 34.8  & 33.22 (±2.8)\\
        Dynamic-Opt. (Alg. \ref{algo2}) & 35.39$\uparrow$ & 33.21$\uparrow$(±2.7) \\
        Inst-Spec-Opt.(\(\mu\)) (Alg. \ref{algo3}) & 35.65$\uparrow$ & 34.57$\uparrow$ (±2.3)\\
        Inst-Spec-Opt.(\(med\)) (Alg. \ref{algo3}) & 35.1$\uparrow$ & 33.81$\uparrow$ (±2.6)\\
        \bottomrule
      \end{tabular}
    }
  \end{center}
  \caption{PSNR values obtained by different methods on SIDD \cite{Abdelhamed2018DenoisingDataset} and CBSD68 \cite{MartinFTM01} datasets. \Cref{algo1}, \Cref{algo2}, and \Cref{algo3} all use UNETR as the surrogate model; \(\mu\) and \(med\) denote mean and median instance-specific optimizations, respectively.}
  \label{tab:results-psnr}
\end{table}

        

The performance metrics against other baseline are shown in \Cref{tab:results-psnr} and \Cref{tab:ssim_results}. Our method demonstrates superior performance compared to Algorithm \ref{algo1} \cite{TYY19} on both datasets. For SIDD, Algorithm \ref{algo2} achieves PSNR values of 36.31 (train) and 36.12 (test), while random search performs competitively with 36.19 (train) and 36.0 (test). Transitioning from Algorithm \ref{algo2} to Algorithm \ref{algo3} results in a slight improvement, with PSNR values increasing to 36.80 (train) and 36.51 (test). A similar trend is observed on the CBSD68 dataset, where Algorithm \ref{algo3} achieves PSNR values of 35.65 (train) and 34.57 (test). We also experimented with median filtering in addition to standard averaging for Algorithm \ref{algo3}. The observed results were marginal, with PSNR values remaining relatively stable across both techniques. When evaluating SSIM, our method outperforms both Algorithm \ref{algo1} and random search, achieving a value of 0.929. 

Futhermore, using Algorithm \ref{algo3}, we experiment with different batch sizes. The PSNR and SSIM values for SIDD with batch sizes of 1, 2, 3 and 4 are shown in \Cref{table:batch_psnr}. This analysis shows that while a batch size of 1 can induce variance, the results remain consistent across different batch sizes, with only minimal deviations in performance.

One constraint in our approach is related to the noise level of the dataset. While evaluating performance on the CBSD68 dataset at a low noise level of 15, we observed that increasing noise levels make parameter optimization more challenging. This limitation is due to BM3D's inherent lack of optimization for high-noise conditions, which results in suboptimal performance compared to deep learning networks specifically designed for high-noise denoising \cite{Claus_2019_CVPR_Workshops, Vaksman_2021_ICCV, chen2020pre}.


\begin{table}[h]
  \begin{center}
    \small{
      \begin{tabular}{l|c|c}
        \toprule
        \multicolumn{1}{c|}{} & \multicolumn{2}{c}{\textbf{SSIM}} \\ 
        \cmidrule(lr){2-3}
        \textbf{Method} & \textbf{SIDD \cite{Abdelhamed2018DenoisingDataset}} & \textbf{CBSD68 \cite{MartinFTM01}} \\ 
        \midrule
        Proxy-Opt. (Alg. \ref{algo1}) & 0.87 (±0.06) & 0.85 (±0.04) \\
        Random Grid Search & 0.92 (±0.07) & 0.91 (±0.05) \\
        Dynamic-Opt. (Alg. \ref{algo2}) & 0.93$\uparrow$(±0.05) & 0.91$\uparrow$(±0.05) \\
        Inst-Spec-Opt.($\mu$) (Alg. \ref{algo3}) & 0.93$\uparrow$(±0.05)& 0.92$\uparrow$(±0.04) \\
        Inst-Spec-Opt.($med$) (Alg. \ref{algo3}) & 0.92$\uparrow$(±0.05)& 0.92$\uparrow$(±0.04) \\
        \bottomrule
      \end{tabular}
    }
  \end{center}
  \caption{SSIM results on SIDD \cite{Abdelhamed2018DenoisingDataset} and CBSD68 \cite{MartinFTM01} datasets. \Cref{algo1}, \Cref{algo2}, and \Cref{algo3} use UNETR as the surrogate model; \(\mu\) and \(med\) denote mean and median instance-specific optimizations, respectively.}

  \label{tab:ssim_results}
\end{table}

        

\begin{figure}[h]
\centering
\begin{subfigure}[b]{0.33\linewidth}
        \includegraphics[width=\linewidth]{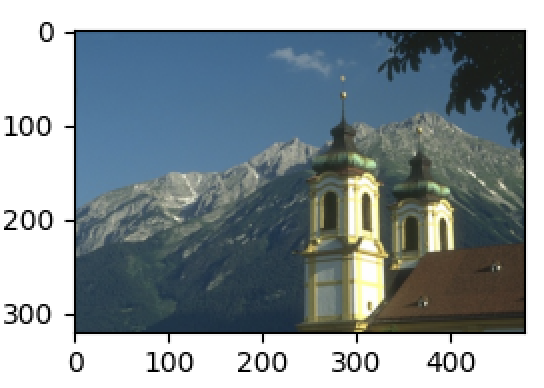}
        \caption{}
        \label{fig:image1}
    \end{subfigure}\hfill
    \begin{subfigure}[b]{0.33\linewidth}
        \includegraphics[width=\linewidth]{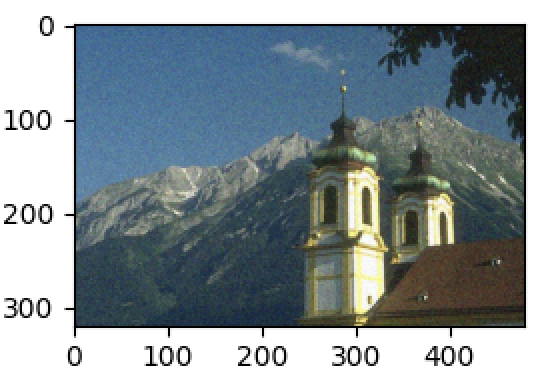}
        \caption{}
        \label{fig:image2}
    \end{subfigure}\hfill
    \begin{subfigure}[b]{0.33\linewidth}
        \includegraphics[width=\linewidth]{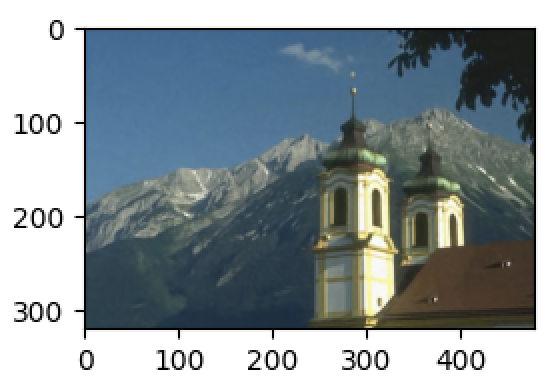}
        \caption{}
        \label{fig:image3}
    \end{subfigure}
    \begin{subfigure}[b]{0.33\linewidth}
        \includegraphics[width=\linewidth]{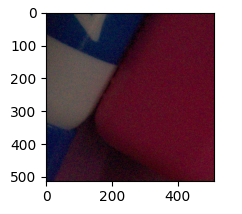}
        \caption{}
        \label{fig:image4}
    \end{subfigure}\hfill
    \begin{subfigure}[b]{0.33\linewidth}
        \includegraphics[width=\linewidth]{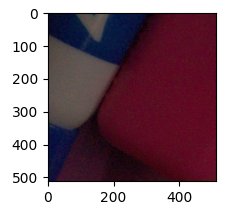}
        \caption{}
        \label{fig:image5}
    \end{subfigure}\hfill
    \begin{subfigure}[b]{0.33\linewidth}
        \includegraphics[width=\linewidth]{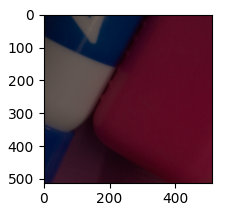}
        \caption{}
        \label{fig:image6}
    \end{subfigure}
    \caption{Visualization of optimization outputs. First row shows sample (a) ground truth (GT), (b) noisy (NOISY), and (c) denoised image produced by Algorithm \ref{algo3} on CBSD68\cite{MartinFTM01}. In second row,  (d) shows the denoised output after employing Algorithm \ref{algo1} \cite{TYY19}, (e) Algorithm \ref{algo2}, and (f) Algorithm \ref{algo3} on the SIDD dataset\cite{Abdelhamed2018DenoisingDataset}.}

    \label{fig:visualization_output}
    \label{fig:onecol}
\end{figure}

\subsection {Optimal Parameters $\theta$}
 For Algorithm \ref{algo1}, the optimal parameters identified for SIDD include a denoising coefficient ($cff$) of 2.8, patch size ($n1$) of 4, color space ($cspace$) set to ``YCbCr", Wiener transform ($wtransform$) set to``dct", and a neighborhood size ($neighborhood$) of 4. For Algorithm \ref{algo2}, the parameters were: $cff$ = 5.3, {$n1$ = 8, $cspace$ = ``opp", $neighborhood$ = 7 and $wtransform$ = ``dct". In the case of Algorithm \ref{algo3}, the parameters are instance-specific, but the patch size ($n1$) was consistently found to be 4, and the transform type ($wtransform$) was found to be ``dct'' (0) in all instances. While most instances (around 80 \%) performed optimally in the ``YCbCr" color space, some instances (around 20\%) showed better performance in the ``opp'' color space. The primary differences in performance can be attributed to the denoising coefficient ($cff$) and the neighborhood size ($neighborhood$). We show the distribution of the parameters values across the test samples of SIDD dataset in Figure \ref{fig:param_dist}. Futhermore, we  show the test set performance based on PSNR for individual images in the scatter plot in \Cref{fig:scatter}, with \Cref{algo3} demonstrating increase in PSNR per instance, marked in blue.
\begin{table}[ht]
    \centering
    \begin{tabular}{|c|c|c|}
        \hline
        \textbf{Batch Size} & \textbf{PSNR (dB)} & \textbf{SSIM} \\
        \hline
        1  & 36.51 (±5.8) & 0.93 (±0.05) \\
        2  & 35.84 (±5.1) & 0.93 (±0.04) \\
        3  & 35.88 (±5.1) & 0.92 (±0.04) \\
        4  & 36.31 (±5.3) & 0.92 (±0.04) \\
        \hline
    \end{tabular}
    \caption{PSNR and SSIM values on the SIDD dataset \cite{Abdelhamed2018DenoisingDataset} test instances, evaluated using Algorithm \ref{algo3} for different batch sizes.}
    \label{table:batch_psnr}
\end{table}


\begin{figure}[h!]
\centering
    \begin{subfigure}[b]{0.8\linewidth}
        \includegraphics[width=\linewidth]{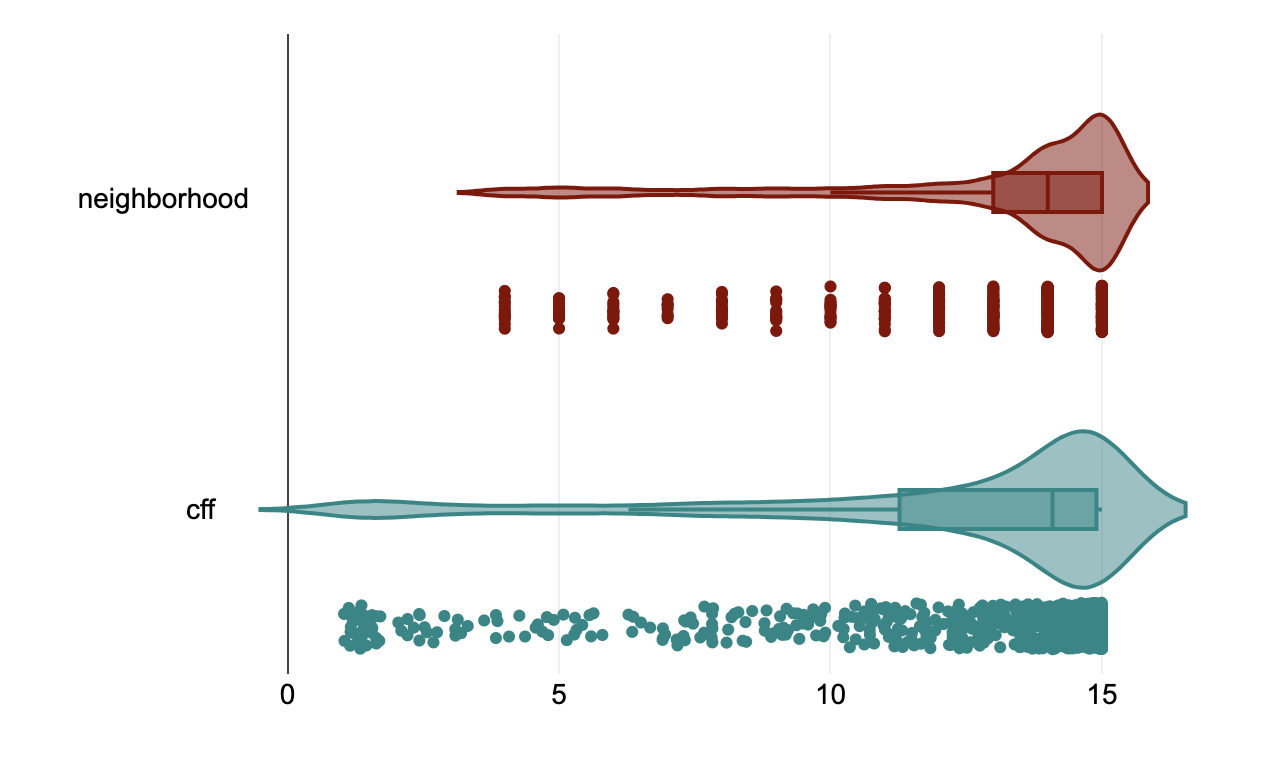}
        \caption{}
        \label{fig:image1}
    \end{subfigure}\vfill
    \begin{subfigure}[b]{0.8\linewidth}
        \includegraphics[width=\linewidth]{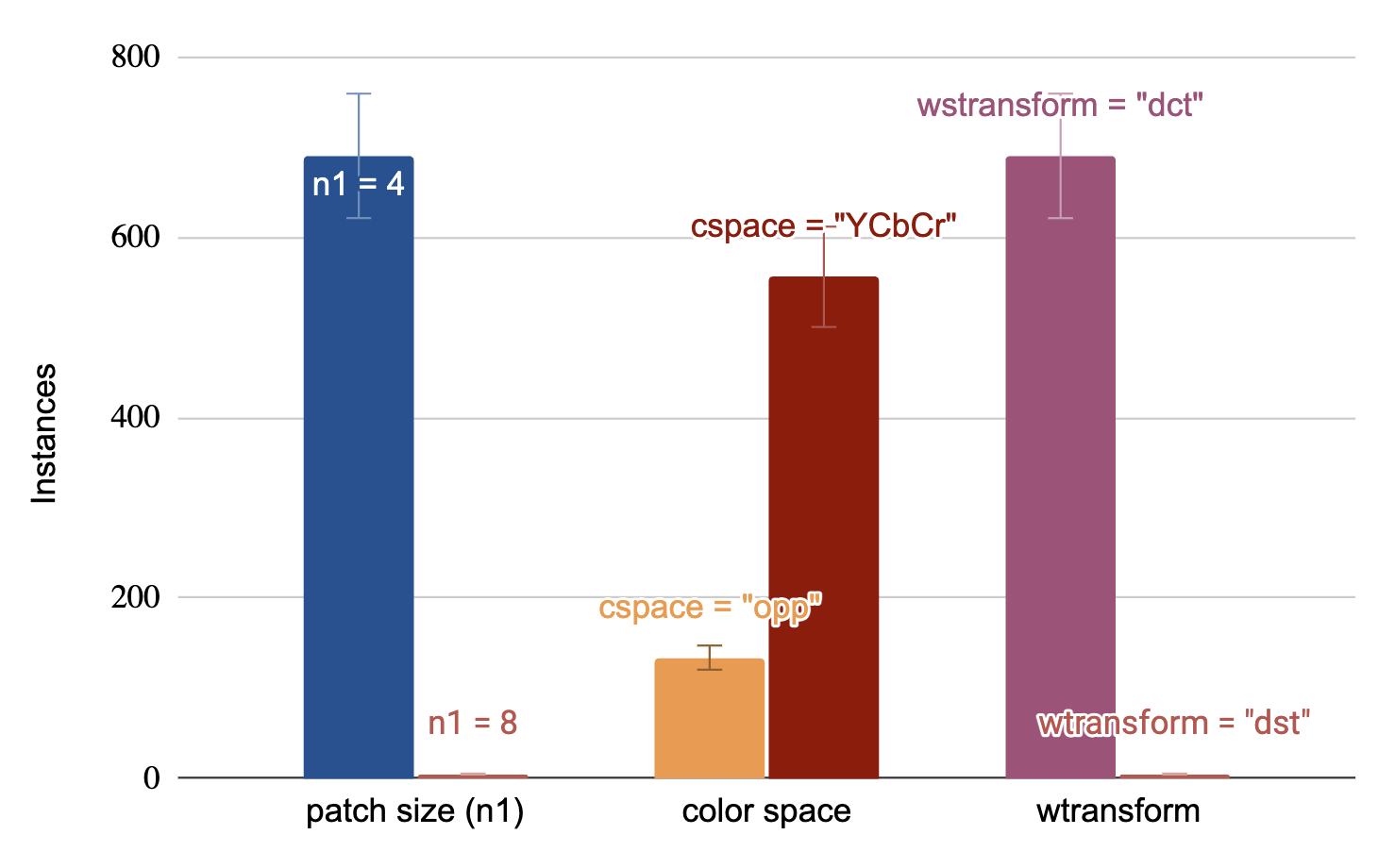}
        \caption{}
        \label{fig:image2}
    \end{subfigure}
    \caption{Optimal parameters learned by the parameter learner, $f_{\text{pl}}$, across all test instances of the SIDD dataset \cite{Abdelhamed2018DenoisingDataset}. (a) displays a violin plot illustrating the distribution of the quantized denoising coefficient ($cff$) and neighborhood size, while (b) presents a histogram showing the frequency of the quantized patch size ($n1$), color space ($cspace$), and Wiener transform ($wtransform$).}
\label{fig:param_dist}
\end{figure}

\section{Conclusion and Future Work}
\label{sec5:conclusion}
In this work, we have introduced a novel approach utilizing surrogate architecture for automated optimization of input-specific parameters in black-box models, focusing on image denoising with the BM3D denoiser. Our method integrates end-to-end training with the black-box model and parameter learning, demonstrating adaptivity and improved performance through dynamic parameter updates.

Moving forward, future research directions include applying our algorithm to diverse applications beyond image denoising. This includes adapting the approach to other types of black-box models and domains, such as natural language processing, reinforcement learning, or even complex systems modeling, where automated parameter optimization can yield substantial benefits. Additionally, efforts will focus on scaling the approach to handle larger datasets and higher-dimensional parameter spaces, aiming to advance automated parameter optimization in complex, real-world systems.










\section*{Acknowledgment}
\label{sec5:ack}
The authors acknowledge the support of Alberta Innovates as the source of funding for this project.
{\small
\bibliographystyle{ieee_fullname}
\bibliography{egbib}
}

\end{document}